\documentclass[sigconf]{acmart}
\usepackage{multirow}
\AtBeginDocument{%
  \providecommand\BibTeX{{%
    \normalfont B\kern-0.5em{\scshape i\kern-0.25em b}\kern-0.8em\TeX}}}

\setcopyright{acmcopyright}
\copyrightyear{2021}
\acmYear{2021}

\acmConference[RAI@KDD '21]{KDD '21}{August 15, 2021}{Singapore}
\acmBooktitle{RAI@KDD '21,
  August 15, 2021, Singapore}

\begin{document}

\title{An Empirical Study of Accuracy, Fairness, Explainability, Distributional Robustness, and Adversarial Robustness}

\author{Moninder Singh}
\affiliation{%
  \institution{IBM Research -- T. J. Watson Research Center}
  \city{Yorktown Heights}
  \state{NY}
  \country{USA}
  \postcode{10598}
}

\author{Gevorg Ghalachyan}
\affiliation{%
  \institution{DataArt}
  \city{Yerevan}
  \country{Armenia}}

\author{Kush R. Varshney}
\affiliation{%
  \institution{IBM Research -- T. J. Watson Research Center}
  \city{Yorktown Heights}
  \state{NY}
  \country{USA}
  \postcode{10598}
}

\author{Reginald E. Bryant}
\affiliation{%
 \institution{IBM Research -- Africa}
 \city{Nairobi}
 \country{Kenya}}

\begin{abstract}
To ensure trust in AI models, it is becoming increasingly apparent that evaluation of models must be extended beyond traditional performance metrics, like accuracy, to other dimensions, such as fairness, explainability, adversarial robustness, and distribution shift. We describe an empirical study to evaluate multiple model types on various metrics along these dimensions on several datasets. Our results show that no particular model type performs well on all dimensions, and demonstrate the kinds of trade-offs involved in selecting models evaluated along multiple dimensions.
\end{abstract}


\maketitle

\section{Introduction}
\label{sec:intro}

Machine learning models are increasingly used to support decision making in high-stakes applications such as finance, healthcare, employment, and criminal justice. In these applications, performance considerations of models that go beyond predictive accuracy are important to ensure their safety and to make them trustworthy \cite{Varshney2019}. Some of these considerations include robustness to distribution shift, fairness with respect to protected groups (defined by attributes such as race and gender), interpretability and explainability, and robustness to adversarial attacks.  

These different desiderata are starting to be recognized as a common theme under the heading of `trustworthy machine learning.' Although this common recognition has started to emerge, it is not obvious how these different performance criteria relate to each other.  Which ones represent trade-offs and which ones can be improved simultaneously? Which kinds of models are better or worse for the different criteria?  How do mitigation or defense techniques affect the different criteria? How consistent are the patterns across varied datasets from different applications? 

While previous works have explored the tradeoffs between accuracy and fairness, or accuracy and robustness, etc.  \cite{kearns, adversarial-accuracy}, to the best of our knowledge, there is no empirical study in the literature that simultaneously evaluates models along various metrics related to all the above-mentioned different pillars of trust for several model types and several datasets. 
In this paper, we address this gap by conducting such an empirical study and reporting multi-dimensional evaluations to start answering the aforementioned questions. 

While the choice/need for evaluation along a particular dimension, as well as the actual metrics used to evaluate models along that dimension, will depend on the actual application (for example, different fairness or performance or explainability metrics may be appropriate for different problem tasks), we initially focus on some common, widely understood metrics for each dimension. In an extended version of the paper, we will expand the analysis to additional metrics along each of the trust pillars considered, as well as expand the choice of models (and hyperparameters) evaluated. Nevertheless, we believe this work represents an important initial step in developing better intuition for the relationships among the different criteria in order to better design overall data mining pipelines that satisfy the requirements of policymakers and other stakeholders. 
Also, the metrics/methodology we compile can be applied to automatically complete certain quantitative portions of AI transparency / documentation proposals such as factsheets and model cards \cite{Arnold2019,Mitchell2019}. 

\section{Multi-dimension Evaluation of AI models}
\subsection{Predictive Performance}

We looked at two measures, accuracy and balanced accuracy. Accuracy (for the binary classifiers) focuses on the total number of correct predictions, and is defined as: 
\[ \frac{TP + TN}{TP + FP + TN + FN} \text{.}\]
where TP is number of True Positives, TN is number of True Negatives, FP is number of False Positives, and FN is number of False Negatives.

Balanced accuracy, on the other hand, is a proportioned measure of accuracy and is used to score models that predict outcomes from imbalanced data. Balanced accuracy is defined as  
\[ \frac{1}{2} \left( \frac{TP}{TP + FP}  + \frac{TN}{TN + FN} \right) \]

Various other performance metrics are also often used, depending upon the task at hand, such as F1 score, AUC-ROC, and True Positive Rate (aka sensitivity). As discussed previously, we intend to report on those metrics (as well as alternative metrics of the other trustworthy pillars discussed below) in an extended version of this paper.

\subsection{Fairness and Bias}
There is a popular notion for fairness which can be numerically quantified: Group versus Individual fairness \cite{dwork2012fairness}. Group fairness analysis involves the examination of the relationship between different groups along so-called sensitive/protected attributes and measures the differences or ratios of desired outcomes. Outcomes can be measured from the data itself or, in our case, from model predictions. Individual fairness examines the rate of outcomes of similar groups of individuals clustered along a number of the attributes, typically excluding sensitive/protected attributes. 
For this work, we considered a commonly used ratio group fairness metric, Disparate Impact \cite{disparate_impact}: 
\[ \frac{Pr(Y = 1 | D = 
\text{unprivileged})}
           {Pr(Y = 1 | D = \text{privileged})} \text{,}\] where $Y = 1$ represents positive or desirable outcomes. A value of 1 implies fairness, as measured by this metric. As previously mentioned, however, other fairness metrics, such as equalized odds or equality of opportunity \cite{equalized-odds} may be more appropriate for some applications/domains than disparate impact, and will be reported in subsequent work.
           
\subsection{Bias Mitigation}
We also explored the effect of performing bias mitigation, i.e. reducing the model unfairness, on the various metrics. We considered a popular pre-processing method, Reweighing \cite{Kamiran2012}, that modifies sample weights for each (privileged/unprivileged group, label) combination differently to reduce unfairness.

\subsection{Explainability/Interpretability}
 
To measure the explainability/interpretability of a model, we generated local explanations for a random data sample using LIME \cite{lime}, and calculated the average faithfulness of the generated explanations. Faithfulness for a data sample was measured as the correlation between the importance assigned by LIME to various attributes in making a model prediction for that sample and the effect of each of the attributes on the confidence of that prediction \cite{faithfulness}.

\subsection{Adversarial Robustness}

From the perspective of ethical AI, stakeholders want to achieve a model that is maximally robust to adversarial attacks \cite{goodfellow2015explaining} wherein insignificant changes made to original features lead to change of model outcome. 
The adversarial robustness of a model is not a generic metric; it is estimated with respect to specific adversarial examples. Here we generated adversarial examples using the model-agnostic HopSkipJump algorithm \cite{hopskipjump} which were then used to evaluate the Empirical Robustness \cite{empirical-robustness}, equivalent to computing the average minimal perturbation that an attacker must introduce for a successful attack.

\subsection{Distribution Shift}
Finally, to evaluate the robustness of various models to distribution shift, we (a priori) created shifted datasets for each evaluated dataset by partitioning each into two parts, based on some attribute (such as region), as described in Section \ref{datasets}.

\section{Datasets}
\label{datasets}

Eight diverse datasets (in terms of size, application, and source) were considered, as described below.

\subsection{Home Mortgage Disclosure Act (HMDA)}
This data contains the 2018 mortgage application data collected in the U.S. under the Home Mortgage Disclosure Act\cite{hmda}. The dataset consists of 1,119,631 applications for single-family, principal-residence purchases. The target is to predict whether a mortgage is approved, and race-ethnicity was treated as the sensitive feature (non-Hispanic Whites/non-Hispanic Blacks). For the purpose of the experiments in this paper, we did not consider people of other races/ethnicity and filtered them out of the data. State was used to partition the data into the base and shift datasets - the 11 Northeastern states (from DE to ME) constituted the shift dataset. Of the 929825 applications in the base dataset, 93.8\% belonged to White applicants and 93.6\% were approved (approval rate for Whites was 94.1\% and Blacks was 85.7\%). Of the 189806 applicants in the shift dataset, 92.7\% belonged to White applicants and 93.8\% were approved (White approval rate was 94.4\% while Black rate was 86.4\%).

\subsection{Mexico Poverty}
Extracted from the Mexican household survey (2016),  this dataset \cite{mexico-dataset} contains demographic and poverty-level features on 70305 Mexican households. The target is predict whether a family is poor or not (`outcome' = 1). For sensitive feature, we consider age (as defined by the `young' feature). We use the `urban' feature to partition the data into the base and shift datasets: urban residents (`urban' = 1) are used for training/testing the model (44672 records) while rural (`urban' = 0) residents are used exclusively for distribution shift testing (25633 records).  There are 53.2\% young families in the base dataset, and 52.5\% in the shift dataset. Finally,
34.9\% of the families are classified as poor in the base dataset (39.5\% of young and 29.7\% of old), while 36.6\% are classified as poor in the shift dataset (41.4\% of young vs. 31.3\% of old).

\subsection{Adult}
The Adult dataset \cite{adult-dataset}, from the UCI ML repository \cite{uci}, consists of demographic and financial data on 48842 individuals extracted from the US Census Bureau database. The aim is to predict whether an individual's income exceeds \$50K/year. We treat `race' as the sensitive feature. The data is partitioned on the feature `native-country': those in  the `United-States' are retained in the base training dataset while the rest are used to create the data for distribution shift. 3620 records with at least one null value are dropped. Consequently, the training dataset consists of 41292 records while the shift dataset has 3930. `Non-whites' constitute 11.98$\%$ of the training dataset and 34.96\% of the shift dataset. The target is true for 25.3\% (26.9\% for `whites', 13.9\% for 'non-whites') of the training dataset and 19.3\% (17.4\% for `whites', 23\% for `non-whites') of the shift data.

\subsection{Bank Marketing}
The Bank Marketing dataset \cite{bank-dataset,uci} from the UCI repository contains data about marketing campaigns made by a Portuguese bank. The task is to predict which people will subscribe to a term deposit. We partition the data into train and shift based on the month in which contact was last made with the customer: if contact was made between September and December, then the person is included in the shift dataset (4790 records), while people with last contact in the remaining 8 months are kept in the training dataset (25698 records). 10700 records were discarded as they had null value fields. The protected attribute is age ($\ge$ 25 or not). In the training dataset, 97\% of the people are older than 25, while the corresponding number for the shift dataset is 98\%. 11.33\% of the people in the training dataset subscribed to a term deposit (18.75\% of the young people and 11.1\% of the old) while 19.79\% of the people in the shift dataset subscribed (57.3\% of the young and 19.02\% of the old).

\subsection{Financial Inclusion in Africa (FIA)}
The FIA dataset \cite{zindi-dataset} contains demographic and financial services data for 33610 individuals across 4 East African countries: Kenya, Rwanda, Tanzania, and Uganda. We used the public `train' component consisting of data on 23524 individuals. The aim is to predict who is likely to own a bank account. The sensitive feature considered is `gender\_of\_respondent', while country is used  to create the dataset to test for distribution-shift: people in Uganda constitute the shift dataset (2101 records), while people from the other three countries constitute the base training dataset (21423 records). The `gender\_of\_respondent', as recorded, is binary valued - male/female. There are 41.7\% males in the training dataset and 34.1\% males in the shift dataset. Finally, 14.6\% have bank accounts in the training dataset (19.5\% of men, 11.1\% of women), while the corresponding number for the shift dataset is 8.6\% (11.7\% of men, 7\% of women).

\subsection{Medical Expenditure Panel Survey (MEPS)}
The Medical Expenditure Panel Survey (MEPS) dataset \cite{meps} is a collection of surveys collected annually by the US Department of Health and Human Services. Each year, a new cohort (called panel) is started and interviewed over five rounds over the next two calendar years. The dataset used \cite{singh2019} consists of 2-year longitudinal data for the cohorts initiated in 2014 (panel 19) as the base training/testing dataset (8136 records) and 2015 (panel 20) as the shift dataset (8737 records).
The objective is to predict patients who would incur high second-year expenditure, based on first-year health and demographic attributes. Race is used as the sensitive feature (64.1\% White in training data and 67.9\% White in shift data). High expenditure patients make up 9\% of people in the training data (10.27\% of Whites and 6.79\% of Blacks), and 10\% in shift dataset (11.37\% of Whites and 7.2\% of Blacks).

\subsection{German Credit}
The UCI German Credit dataset \cite{uci} describes the creditworthiness of 1000 people. The task is to predict which people have good credit. We use `sex' as the sensitive attribute. The dataset is split based on the `foreign worker' attribute: foreign workers are retained in the base training data (963 records), while the non-foreign workers constitute the shift dataset (37 records). There are 68.54\% males in the training dataset and 81.10\% males in the shift dataset. In the training dataset, 69.26\% have good credit (71.37\% of males, 64.69\% of females) while 89.19\% have good credit in the shift dataset (93.3\% of males, 71.4\% of females).

\subsection{Heart Disease}
The Cleveland Heart dataset \cite{heart-dataset,uci} from the UCI repository is a small 303 case dataset (from which 6 records are discarded for null values). Due to the size of the dataset, we do not create a distribution-shift dataset. The sensitive feature was age above/below the average value (54.5 years): 53.5\% have age above this value. The goal is predict the 'presence of heart disease' (46.1\%).
59.75\% have heart disease among patients who have age above mean, while 30.43\% have heart disease for age under mean.

\section{Experiments}
We tested four classification algorithms: gradient boosting (GBC), random forests (RF), logistic regression (LR), and multilayer perceptron (MLP). 5-fold cross validation was used for the experiments. For each cross-validation split, the following was done:
\begin{enumerate}
\item The training split was used to build the four models. Categorical features were one-hot encoded, and feature standardization was done by centering 
and scaling 
The trained models were then evaluated on the test split.
\item Each model was further tested on the shifted dataset.
\item Bias mitigation was performed on the training split and the debiased data was used to build the four models which were once again tested on the test splits.
\item The models learned above from the debiased data were also tested on the shifted dataset.
\end{enumerate}
The means and standard errors for all the metrics were calculated from the five splits. For computing faithfulness of explanations, LIME \cite{lime} was used to generate local explanations and a random sample of 50 records was used to compute the mean faithfulness score. For measuring empirical robustness, a random sample of 20 records was used for generating adversarial samples in each run.
Three open-source toolkits, AI Fairness 360 \cite{aif360}, AI Explainability 360 \cite{aix360}, and the Adversarial Robustness Toolbox \cite{nicolae2019adversarial}, were used for evaluating model fairness/performing bias mitigation, evaluating explainability, and measuring model adversarial robustness, respectively.

\begin{figure}[ht]
  \centering
  \includegraphics[width= \linewidth]{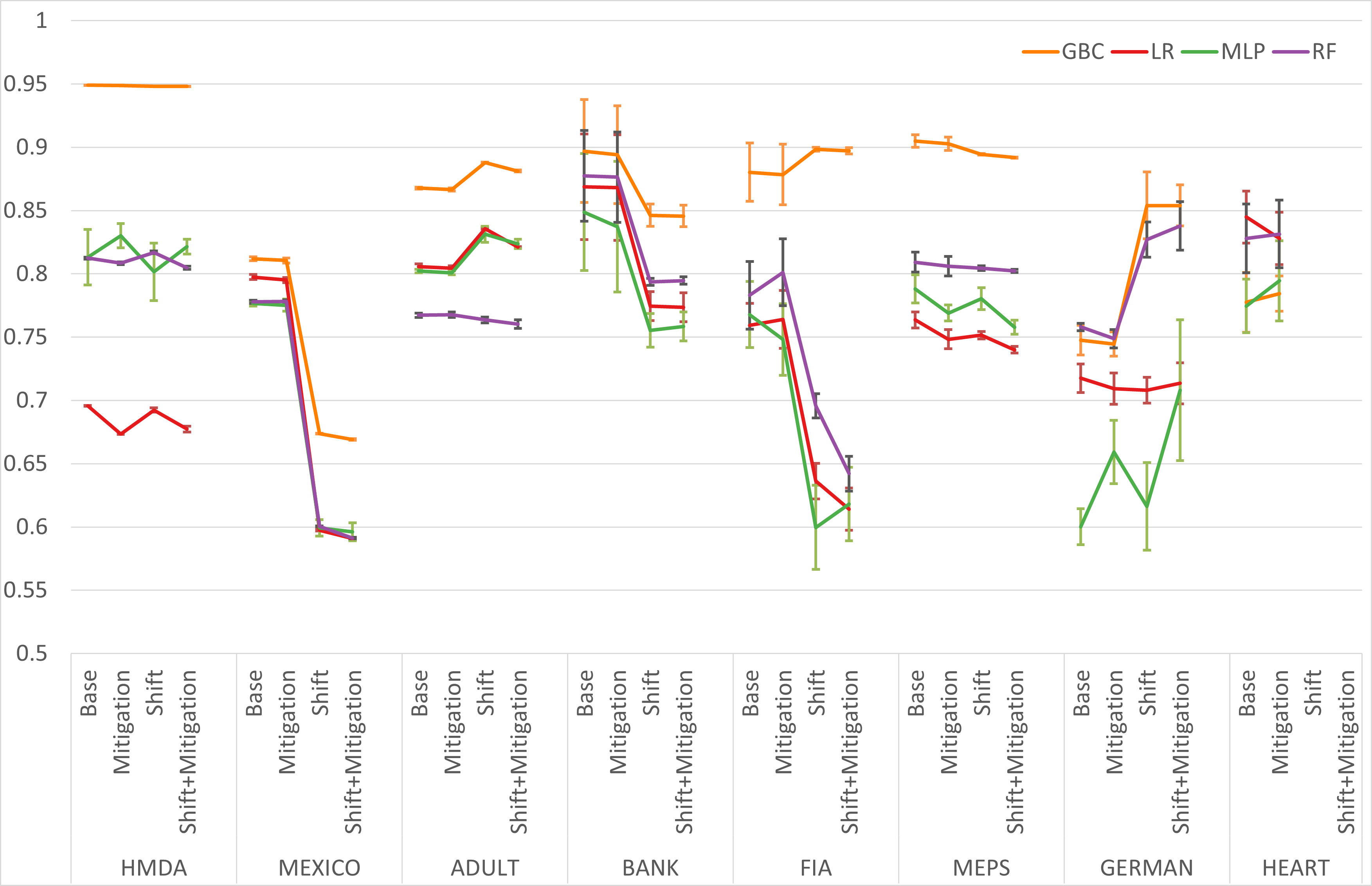}
  \caption{Accuracy.}
  \label{fig:accuracy}
  \Description{Graphs showing accuracy results for various models/datasets..}
\end{figure}

\begin{figure}[ht]
  \centering
  \includegraphics[width=\linewidth]{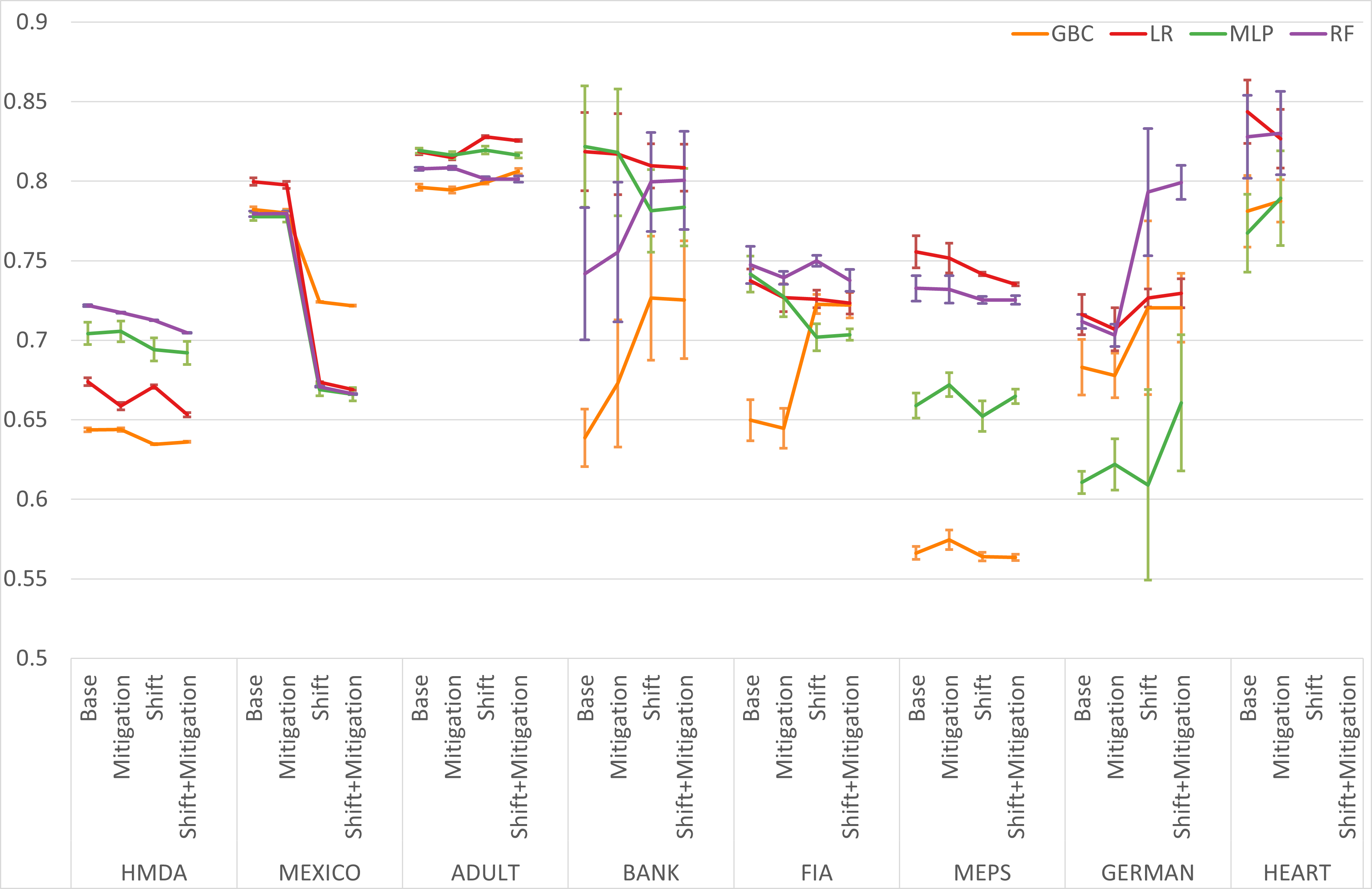}
  \caption{Balanced accuracy.}
  \label{fig:balanced_accuracy}
  \Description{Graphs showing balanced accuracy results for various models/datasets..}
\end{figure}

\begin{figure}[ht]
  \centering
  \includegraphics[width=\linewidth]{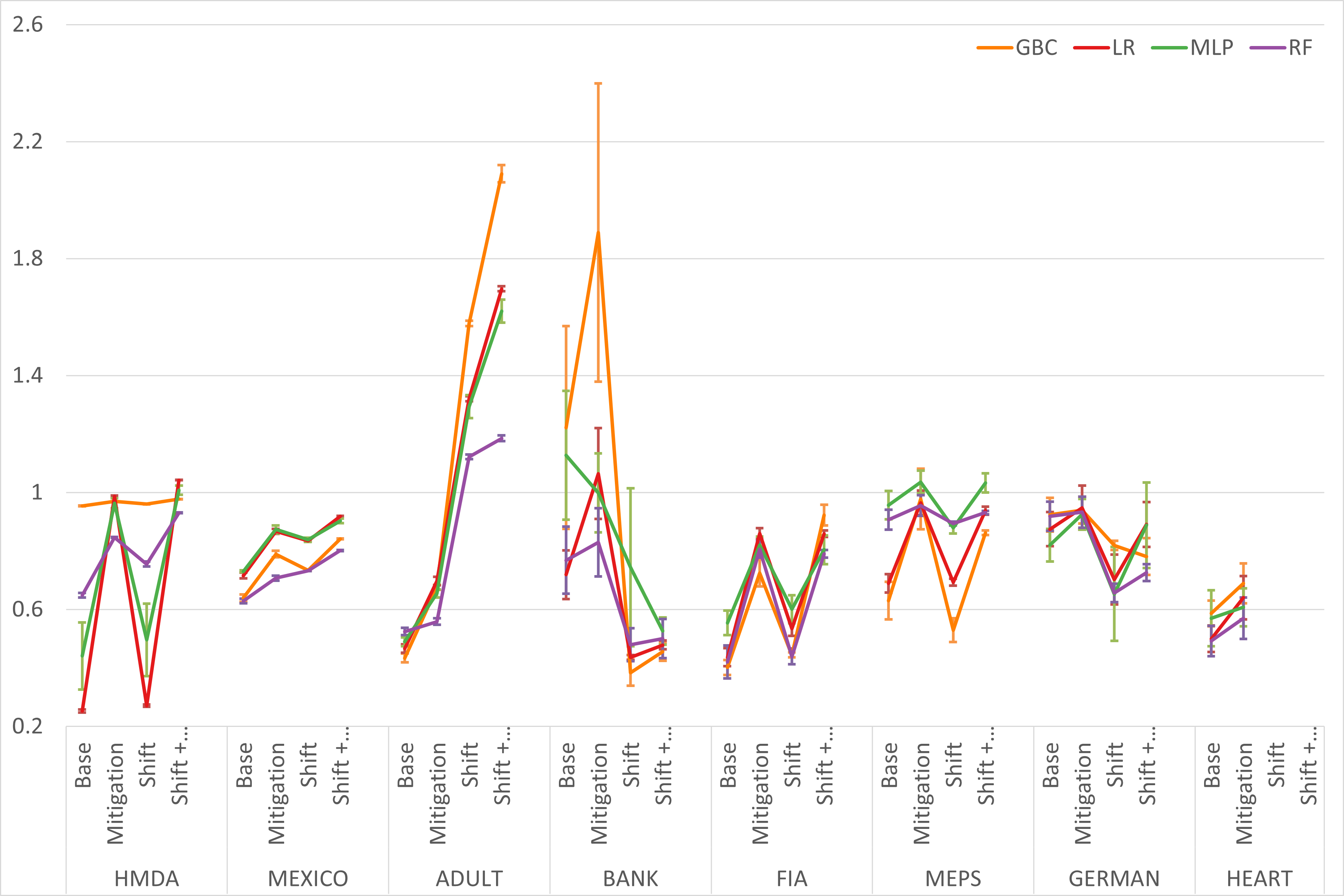}
  \caption{Disparate Impact.}
  \label{fig:fairness}
  \Description{Graphs showing fairness metric results for various models/datasets..}
\end{figure}

\begin{figure}[ht]
  \centering
  \includegraphics[width=\linewidth]{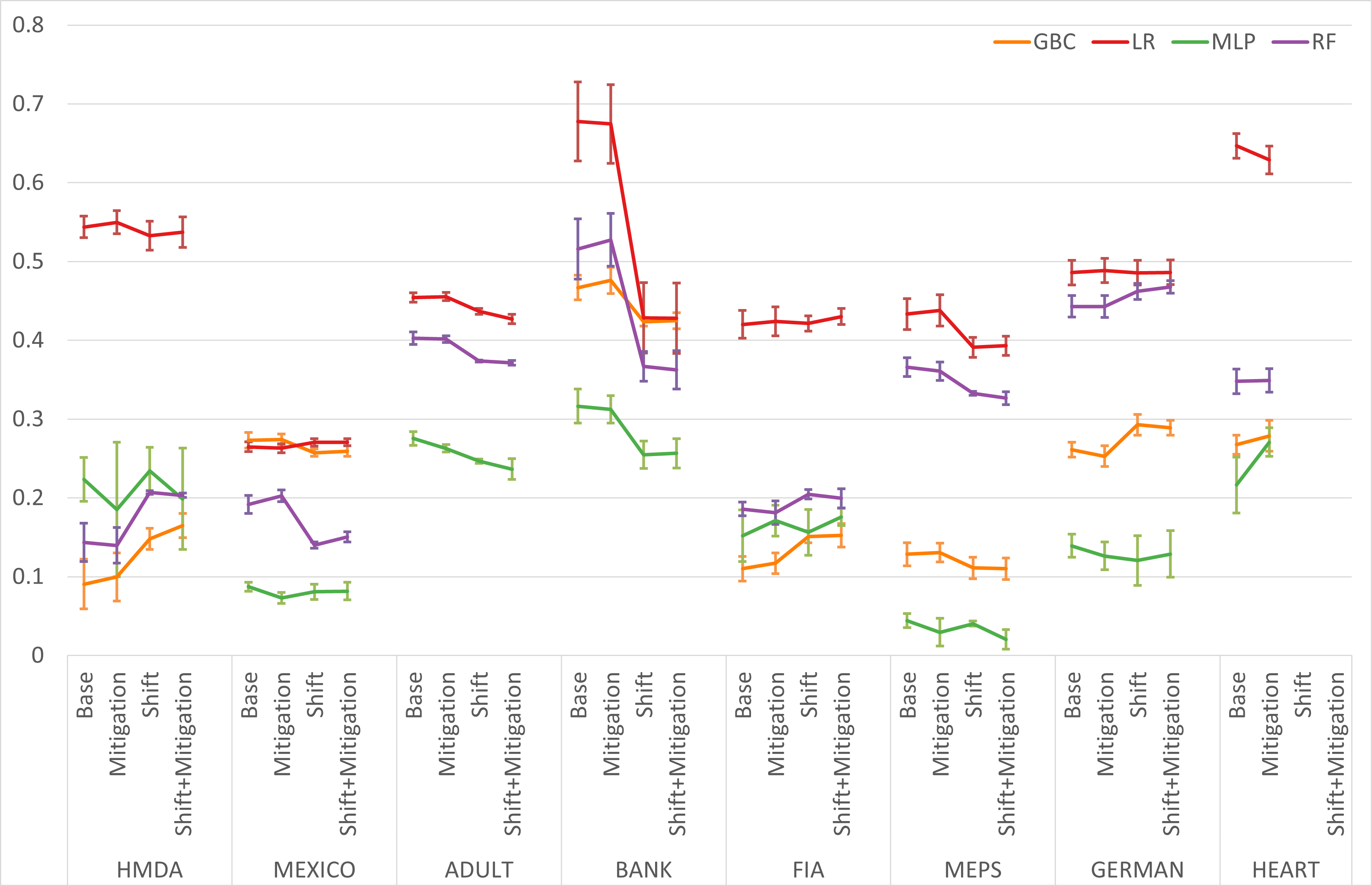}
  \caption{Faithfulness.}
  \label{fig:faithfulness}
  \Description{Graphs showing faithfulness metric results for various models/datasets.}
\end{figure}

\begin{figure}[ht]
  \centering
  \includegraphics[width=\linewidth]{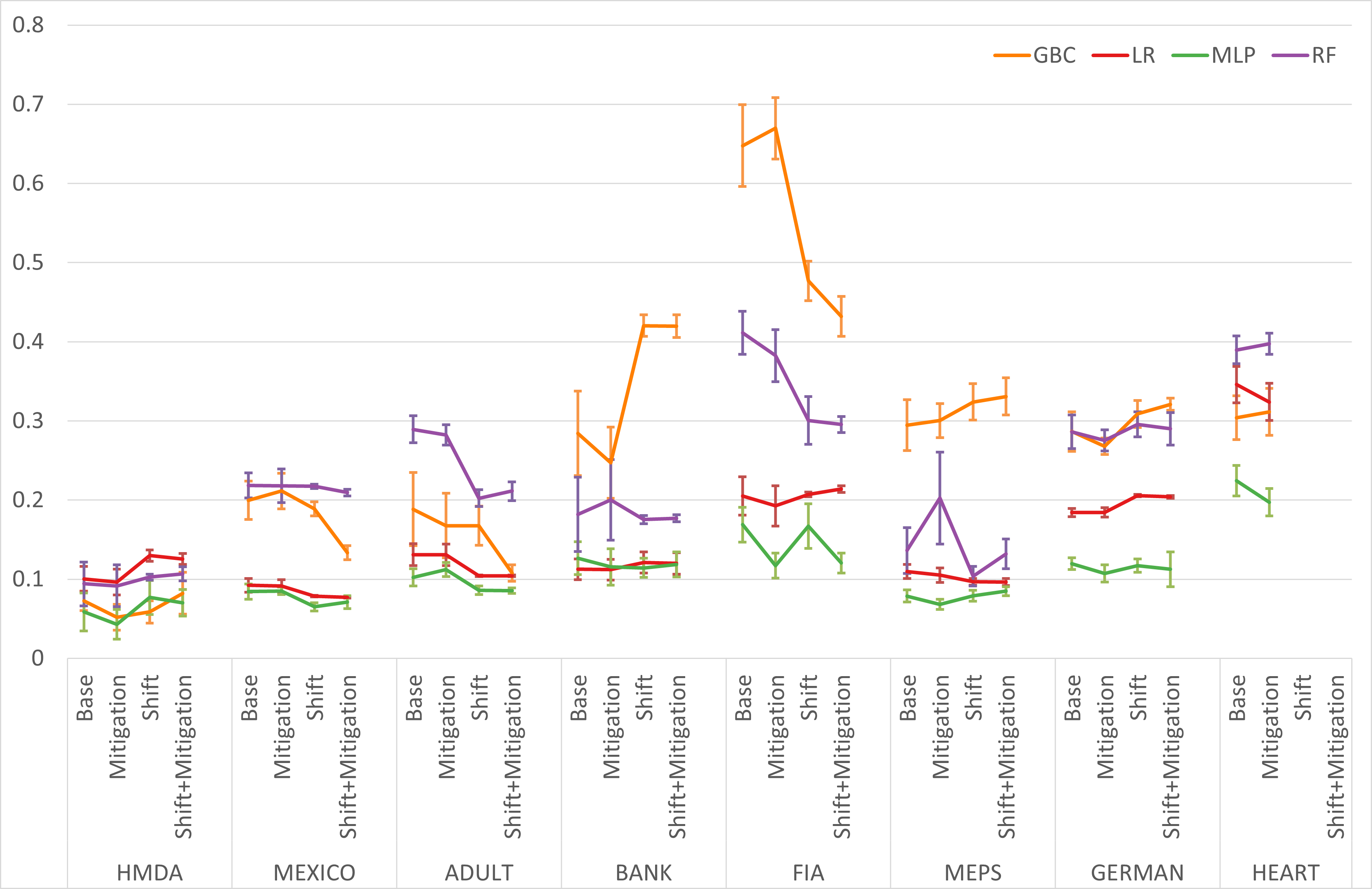}
  \caption{Empirical Robustness.}
  \label{fig:robustness}
  \Description{Graphs showing adversarial robustness metric results for various models/datasets.}
\end{figure}

\section{Results and Discussion}

Fig.\ \ref{fig:accuracy} and \ref{fig:balanced_accuracy} show how models learned with different algorithms performed on the various datasets in terms of accuracy and balanced accuracy. While GBC models typically had the best accuracy across the datasets (7 of 8), they typically were amongst the worst performers when it came to balanced accuracy. 
Meanwhile, MLP models typically performed poorly in terms of accuracy, but exhibited relatively better balanced accuracy, chiefly on the larger datasets.
RF models, although often not the best, performed well on most of the datasets on all the metrics, especially the smaller datasets. LR models similarly often showed good balanced accuracy, but relatively poor accuracy.

With regards to fairness, most of the datasets have substantial disparity between the privileged/unprivileged groups as defined by the protected features used. However, this unfairness did not manifest itself the same way in all the learned models (Fig.\ \ref{fig:fairness}).
For example, for the HMDA dataset, the LR model was quite unfair with DI of 0.25 but the GBC model was almost fair at 0.95. In the case of the Bank dataset, two of the models were biased against young people but the other two were highly biased against older ones. However, for some of the other datasets, such as Adult and Mexico, the degree of unfairness was similar for all the models.

In terms of the quality of local explanations generated by LIME (Fig.\ \ref{fig:faithfulness}), LR models typically performed the best, while MLP models were generally the worst.
GBC models also exhibited poor faithfulness (often amongst the bottom two), while the RF models were relatively better (typically second best).

Turning to robustness to adversarial attacks (Fig.\ \ref{fig:robustness}), GBC and RF models were typically the most robust, while LR and MLP performed relatively poorly, with the MLP model often being the worst.

In all cases, bias was substantially reduced by mitigation using reweighing, as seen by the peaks in the 2nd of the line graphs. Though the extent of fairness achieved was dependent upon  the extent of original (pre-mitigation) unfairness, RF models were typically not able to benefit from bias mitigation pre-processing as much as GBC, LR, and MLP models (as evidenced by the slope of the lines between the first two points).
While bias mitigation did tend to reduce accuracy of the models, the decrease was fairly small for GBC and RF models but often more pronounced for MLP models.
Bias mitigation also typically tended to yield a small change in explainability (faithfulness). It often reduced faithfulness for MLP models but had little to no effect for LR and RF models
Similarly, adversarial robustness did tend to decrease through bias mitigation, sometimes quite markedly, though there were a few instances where it showed an increase.

Accuracy tended to suffer under distribution shift (3rd point in each line graph). The extent of the drop varied, though it was substantial (10-15\%) for some models and datasets (Mexico, FIA, and Bank). GBC models accuracy was quite robust to distribution shift with typically much smaller changes compared to the other models. 
Explainability (faithfulness) similarly tended to undergo a small decline under distribution shift, except for the Bank dataset where the decrease was substantial, especially in the case of the best performing LR model.
Adversarial robustness suffered quite a lot under distribution shift, especially for GBC and RF models. Fairness similarly suffered a lot under distribution shift.

Trends similar to those seen for bias mitigation in the base case were observed in the distribution shift setting as well (4th points in the line charts). Bias mitigation tended to yield a small drop in accuracy, robustness, and faithfulness. The performance of different types of models on the various metrics was similar to what was observed in the base case.

These results further reinforce the point that models should be evaluated across multiple dimensions and metrics rather than just a single metric, such as accuracy.

\bibliographystyle{ACM-Reference-Format}
\bibliography{sample-base}

\end{document}